\documentclass{article}

\usepackage{arxiv}

\usepackage[utf8]{inputenc} % allow utf-8 input
\usepackage[T1]{fontenc}    % use 8-bit T1 fonts
\usepackage{hyperref}       % hyperlinks
\usepackage{url}            % simple URL typesetting
\usepackage{booktabs}       % professional-quality tables
\usepackage{amsfonts}       % blackboard math symbols
\usepackage{nicefrac}       % compact symbols for 1/2, etc.
\usepackage{microtype}      % microtypography
\usepackage{lipsum}
\usepackage{graphicx}
\graphicspath{ {./images/} }

\usepackage{float}
\usepackage{subfigure}
\usepackage{multirow}

\usepackage{array}

\title{Document- Level Event Extraction with Definition-Driven ICL}

\author{
 Zhuoyuan Liu \\
 School of Information Science \\
  Beijing Language and Culture University\\
  \texttt{202321198862@stu.blcu.edu.cn} \\
   \And
 Yilin Luo \\
  School of Information Science \\
  Beijing Language and Culture University\\
  \texttt{202321198857@stu.blcu.edu.cn} \\
 \\
}

\begin{document}

\maketitle
\begin{abstract}
In the field of Natural Language Processing (NLP), large language models (LLMs) show immense potential in document-level event extraction tasks. However, existing methods face challenges in prompt design. Addressing this issue, we propose an optimization strategy called "Definition-driven Document-level Event Extraction (DDEE)." By adjusting prompt length and enhancing heuristic clarity, we significantly enhance LLMs' performance in event extraction. We employ data balancing techniques to mitigate the long-tail effect, improving the model's generalization in recognizing event types. Additionally, we refine prompts to ensure they are both concise and comprehensive, catering to LLMs' sensitivity to prompt style. Furthermore, the introduction of structured heuristic methods and strict constraints enhances the precision of event and argument role extraction. These strategies not only address prompt engineering issues in LLMs for document-level event extraction but also advance event extraction technology, offering new research perspectives for other tasks in the NLP domain.
\end{abstract}

% keywords can be removed

\section{Introduction}
The field of Natural Language Processing (NLP) has made significant strides in parsing and understanding human language, with event extraction technology playing a crucial role. Event extraction identifies events and their core elements from unstructured text data, detailing participants ("who"), time ("when"), location ("where"), event description ("what"), cause ("why"), and manner ("how"). Accurately extracting this information is essential for tasks such as text summarization, knowledge graph construction, intelligent question answering systems, and recommendation systems.

Event extraction consists primarily of two subtasks: Event Detection (ED) and Event Argument Role Extraction (EAE). Event Detection aims to identify mentioned events in text, while Event Argument Role Extraction further identifies entities involved in the events and their respective roles. Currently, using large-scale pre-trained language models (LLMs) for closed-domain document-level event extraction has become mainstream in the field.

In the field of event extraction in Natural Language Processing (NLP), despite significant advancements, several critical issues remain to be addressed. Firstly, pre-trained and fine-tuned models exhibit shortcomings in generalizing to unseen events, limiting their application in new scenarios. Secondly, existing methods heavily rely on high-quality annotated data, leading to high annotation costs and challenges in data-scarce domains. Additionally, error propagation during event extraction can cause cascading effects, impacting result accuracy. The complex syntactic structures in Chinese sentences, phenomena of arguments spanning multiple sentences, co-occurrence of multiple events, and issues with trigger word absence further exacerbate extraction challenges. Furthermore, instability in contextual learning (ICL) strategies, influenced by various factors, restricts predictive performance of models.

However, pre-trained and fine-tuned models often fall short in generalization performance, and high annotation costs and risks of error propagation are significant constraints to their development. Document-level event extraction faces notable challenges: in English domains, scarcity of high-quality datasets and insufficient generalization of models to unseen events are major bottlenecks; whereas in Chinese domains, challenges include complex sentence structures, argument spanning across sentences, coexistence of multiple events in documents, and trigger word absence, often leading to neglect of prior information.

The application of large-scale language models is heavily constrained by prompt design, with models like ChatGPT showing sensitivity to prompts especially in complex tasks, necessitating further optimization. To address these issues, researchers are exploring strategies such as augmenting document-level datasets, employing graph and semantic analysis models to reduce dependency on predefined events, and minimizing the need for data annotation through prompt engineering, thereby avoiding complex fine-tuning processes.

Moreover, Incremental Context Learning (ICL) as a strategy for task execution based on contextual information exhibits high variability in practical applications. Model predictions are influenced by factors such as example sequence, temporal relevance of labels, prompt formats, and training data distribution. To enhance performance, researchers are experimenting with methods such as self-assessment, expert validation, and emphasizing the importance of effective Chain of Thought (CoT) reasoning prompts, which should directly relate to reasoning tasks, express diversity, decompose complex tasks, integrate known facts for reasoning, and progressively refine and revise reasoning steps.

In this paper, we propose a series of innovative contributions aimed at advancing event extraction technology and addressing the limitations of existing methods:

     \paragraph{Prompt Length Optimization.}For document-level event extraction tasks, we finely tuned the prompt length to optimize overall extraction performance. This adjustment ensures that the prompt contains sufficient information while avoiding efficiency degradation caused by excessive length.

     \paragraph{Task Decomposition and Historical Information Utilization.}We decomposed the event extraction task into two key steps — event detection and trigger word extraction, as well as argument identification and classification. By integrating historical information into these steps, we effectively mitigate error propagation and enhance task accuracy.

      \paragraph{Enhanced Prompt Heuristic Methods.}We refined the prompt heuristic methods, providing a structurally complete and clearly defined framework for events and arguments, incorporating necessary constraints. This not only improves the precision of closed-domain extraction but also provides the model with richer and more consistent reference benchmarks.

     \paragraph{Chain of Thought Approach.}We adopted a chain of thought approach, guiding the model through incremental reasoning by presenting coherent examples. These examples demonstrate how to decompose complex problems into more manageable sub-problems and enhance the model's reasoning capabilities by simulating human thought processes.

      \paragraph{Data Balancing Techniques.}To mitigate the impact of long-tail effects on model performance, we employed a combination of undersampling and oversampling data balancing techniques. This approach ensures representative event type distributions in the dataset, thereby enhancing model generalization.

The aim of this paper is to offer new perspectives and approaches through these innovative methods and technologies, aiming to construct more robust and adaptable event extraction models. We believe that these contributions will improve task performance, strengthen the model's ability to grasp complex relationships between events and arguments, and contribute to further breakthroughs in the field of natural language processing.

\section{Background}
\label{sec:headings}
\paragraph{The application of large language models in event extraction tasks.}
Current research indicates that while large language models perform well in simple event extraction tasks, their robustness and accuracy still need improvement when faced with complex and long-tail scenarios. To address the diverse and complex application requirements in real-world settings, researchers are focusing on optimizing model performance through well-designed prompts, multi-turn dialogue approaches, and integration of role knowledge.

The use of large language models (LLMs) in natural language processing (NLP) tasks has significantly increased, especially with closed models like PaLM \cite{1} , Claude \cite{2} , and GPT-4 \cite{3}. However, despite their satisfactory performance in simple scenarios, their robustness and accuracy in complex tasks still require enhancement. Jun Gao et al. \cite{4}evaluated the performance of large language models in event extraction tasks and found that they perform poorly in complex and long-tail scenarios, particularly when positive samples are removed and event type definitions are unclear. In Chinese event extraction tasks, ChatGPT still faces challenges such as segmentation ambiguity, nested entity recognition, and domain-specific adaptability. Bao Tong et al. \cite{5} suggest optimizing the performance of large language models through well-designed prompts and multi-turn dialogues, while avoiding overly subjective or domain-specific queries to improve output accuracy.

In the zero-shot event extraction domain, Zhigang Kan et al. \cite{6} improved argument recognition performance through a multi-turn dialogue approach, demonstrating its potential in event detection. Fatemeh Shiri et al. \cite{7}optimized the application of LLMs in knowledge graph construction and decision support by integrating advanced prompt techniques such as Chain-of-Thought and Retrieval Augmented Generation, reducing hallucination risks and enhancing accuracy. Ruijuan Hu et al. \cite{8} proposed the Role Knowledge Prompting for Document-level Event Argument Extraction (RKDE) method, which significantly improves extraction accuracy through role knowledge guidance and prompt adjustments, validated on RAMS and WIKIEVENTS datasets. This method combines role knowledge with prompts to optimize the argument extraction process and improve result accuracy. Zhang et al. \cite{9} introduced the ULTRA framework, which enhances LLM performance through hierarchical modeling and refinement methods, reducing the need for expensive API calls.

These innovative approaches and techniques aim to provide new perspectives and methodologies for advancing event extraction technology, addressing current methodological limitations effectively.
\paragraph{LLMs' Capability in Contextual Learning (ICL).}
Large language models (LLMs) demonstrate robust adaptability through In-Context Learning (ICL) without fine-tuning on task-specific datasets. Research in this field is rapidly advancing, with Hanzhang Zhou et al. \cite{58} innovatively studying the use of examples to teach LLMs heuristic rules for handling specific tasks. They optimize example selection strategies and significantly enhance model performance on new categories through analogical reasoning prompts, thereby improving efficiency and accuracy in handling complex tasks. Prophet framework proposed by Yu, Zhou et al. \cite{59} integrates answer candidates and answer-aware context examples as heuristic information, markedly boosting performance in knowledge-based Visual Question Answering (VQA) tasks and demonstrating compatibility with various VQA models and LLMs. Additionally, Qian Li et al. \cite{60} develop an event extraction method using reinforcement learning and task-oriented dialogue systems. By clarifying relationships between arguments and optimizing extraction sequences, they enhance the accuracy of event role classification across different textual contexts.

These studies underscore the adaptability and potential of LLMs across diverse linguistic tasks. They highlight that carefully designed prompts and heuristic rules can effectively enhance model performance without the need for fine-tuning on task-specific datasets, pointing towards new directions in the application and development of future LLMs.
\paragraph{Heuristic Learning in Prompt Engineering.}
In-Context Learning (ICL) \cite{61} is a strategy that enables pre-trained language models to quickly adapt to different tasks with minimal \cite{62} or zero-shot \cite{63} data. This approach avoids explicit fine-tuning by allowing models to understand and execute tasks based on contextual information. As a subset, Few-shot learning utilizes limited annotated samples to train models, employing techniques like pattern utilization to enable large language models (LLMs) to effectively learn new tasks and demonstrate generalization capabilities across tasks \cite{64}.

However, ICL exhibits high instability in practical applications, where model predictions are influenced by factors such as example order, input length, prompt format, and training data distribution \cite{65}. Researchers have improved model accuracy and robustness by optimizing example selection, introducing auxiliary information, generating pseudo inputs and examples, using soft-label tagging, incorporating positive and negative samples, and building expert pools.

Weber et al. \cite{66} enhanced model accuracy by employing well-designed efficient prompt templates and diverse prompt formats. Jiang et al. \cite{67} proposed the P-ICL (Point In-Context Learning) framework, providing critical information about entity types and classifications to LLMs, thereby enhancing named entity recognition tasks. Brunet et al. \cite{68} introduced a new approach, ICL Markup, which optimizes performance in contextual learning by using soft-label tagging, reducing arbitrary decisions in task adaptation, and demonstrating improvements across various classification tasks. Mo et al. \cite{69} introduced C-ICL (Contrastive In-context Learning), which enhances LLMs' performance in information extraction tasks by introducing positive and negative samples in context learning.

Chen et al.'s SELF-ICL framework \cite{70}and Yang et al.'s Auto-ICL framework \cite{71} respectively enhance model adaptation capabilities through self-generated pseudo inputs and auto-generated examples. Qu et al.'s DEEP-ICL (Definition-Enriched Experts for Language Model In-Context Learning) method \cite{72} effectively improves ICL performance through five stages: expert pool construction, task definition extraction, guided retrieval, expert integration, and continual few-shot learning. Liu et al. \cite{73} improved ICL's specificity and efficiency with a retrieval-based strategy. Furthermore, researchers have extensively studied the robustness of ICL, particularly examining how prompt template design details \cite{74} and example arrangement \cite{75}impact ICL's stability and effectiveness, both critical factors determining its reliable and effective operation.
\paragraph{Chain-of-Thought (CoT) Reasoning Optimization via Prompts}
In the field of natural language processing (NLP), simulating logical reasoning capabilities is a key focus of research. The Chain-of-Thought (CoT) method guides large language models (LLMs) through detailed reasoning path examples, systematically breaking down problems into smaller sub-problems and solving them step-by-step. Zero-shot CoT \cite{63}, with simple prompts like "Let’s think step by step," enhances the transparency and accuracy of the reasoning process. Effective prompts for reasoning should possess the following characteristics: direct relevance to tasks, diverse expressions, guidance for model problem decomposition, integration of known facts for reasoning, and step-by-step refinement of processes.

Research indicates that explicit prompting methods for LLMs to decompose problems, such as Least-to-Most\cite{76}and zero-shot CoT, improve the reliability of reasoning. Furthermore, the application of various prompting techniques\cite{62,63,77}has confirmed the effectiveness of decomposition strategies, enabling models to systematically handle complex issues.

Recently, Jin et al. \cite{78} proposed the Exploration of Thought (EoT) prompting method, using evolutionary algorithms to dynamically generate diverse prompts, significantly enhancing LLMs' performance in arithmetic, common sense, and symbolic reasoning tasks. Wang et al.'s \cite{79} Plan-and-Solve (PS) prompting method guides models to formulate and execute plans to solve complex problems, improving performance in multi-step reasoning tasks. Zhao et al.'s \cite{80}Logical Thoughts (LoT) framework utilizes principles of symbolic logic to systematically verify and correct reasoning steps, enhancing LLMs' reasoning capabilities across diverse domains. Kim et al.'s \cite{81} fine-tuning dataset COT COLLECTION enhances the generalization ability of small-scale language models on multi-task unseen problems. Wang et al.'s \cite{82} Cue-CoT method introduces intermediate reasoning steps before generating answers, improving LLMs' performance in handling in-depth dialogue issues.

In summary, innovative chain-of-thought methods such as CoT, EoT, PS prompts, and LoT frameworks significantly enhance the performance of large language models in handling complex logical reasoning tasks within the field of natural language processing. These advancements demonstrate the continual progress and wide-ranging application potential of natural language understanding and reasoning capabilities.

% \subsection{Headings: second level}
% \lipsum[5]
% \begin{equation}
% \xi _{ij}(t)=P(x_{t}=i,x_{t+1}=j|y,v,w;\theta)= {\frac {\alpha _{i}(t)a^{w_t}_{ij}\beta _{j}(t+1)b^{v_{t+1}}_{j}(y_{t+1})}{\sum _{i=1}^{N} \sum _{j=1}^{N} \alpha _{i}(t)a^{w_t}_{ij}\beta _{j}(t+1)b^{v_{t+1}}_{j}(y_{t+1})}}
% \end{equation}

% \subsubsection{Headings: third level}
% \lipsum[6]

% \paragraph{Paragraph}
% \lipsum[7]

\section{Approach}

\begin{figure}
    \centering
    \includegraphics[width=0.75\linewidth]{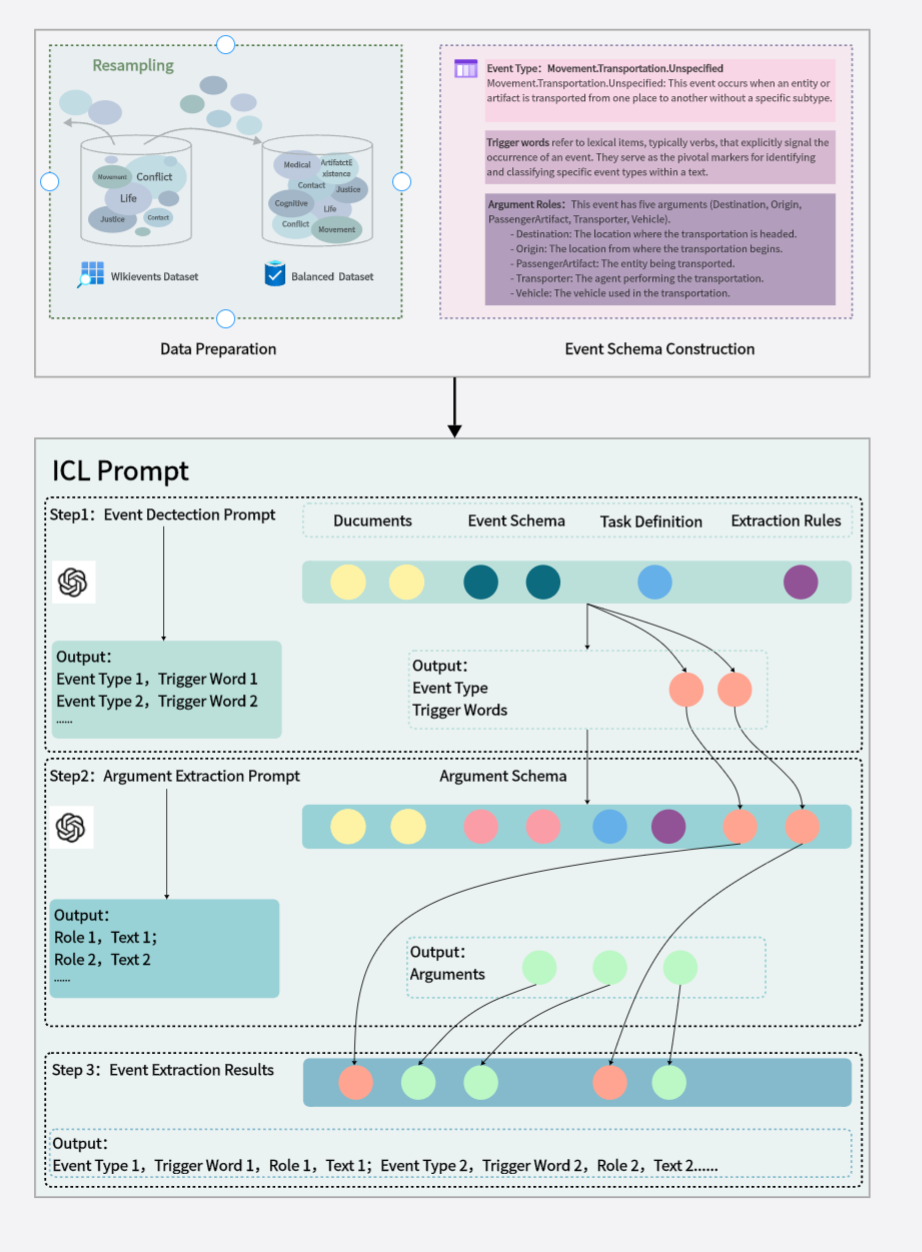}
    \caption{\textbf{Event Extraction Framework.} In this study, we conducted resampling of the Wikievents dataset and clearly defined event types and argument roles. Utilizing a heuristic event framework, we performed a two-step engineering process on the balanced dataset: first extracting event types and triggers, and then applying these results to argument extraction, aiming to enhance the accuracy and efficiency of event extraction.}
    \label{fig:1}
\end{figure}

\begin{figure}
    \centering
    \includegraphics[width=0.75\linewidth]{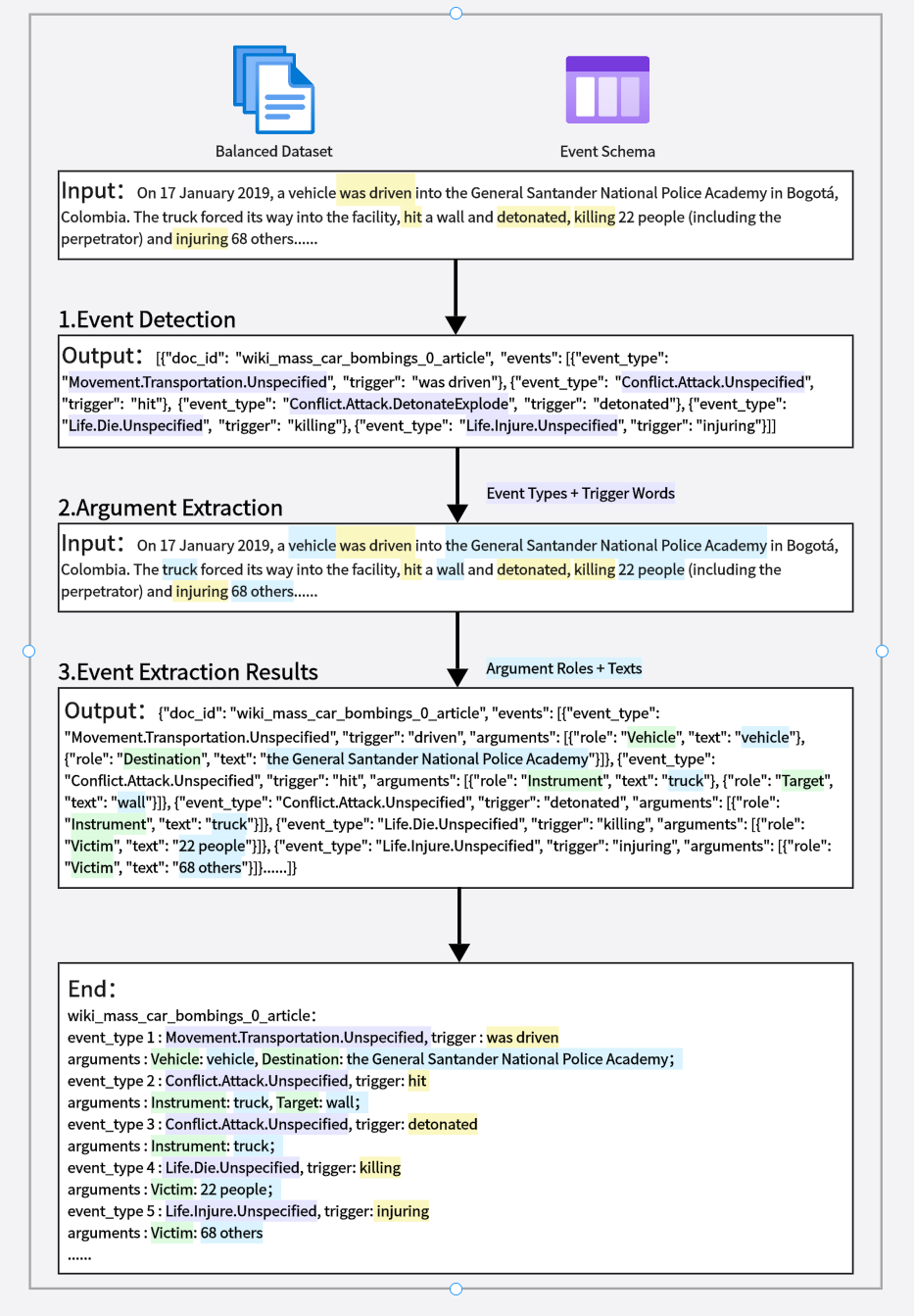}
    \caption{\textbf{Event Extraction Process based on balanced dataset.} This figure provides a detailed illustration of the event extraction process based on a balanced dataset, encompassing event detection, argument extraction, and the final event extraction results. Specifically, it demonstrates how various types of events (such as Movement.Transportation.Unspecified and Conflict.Attack.Unspecified) and their triggers (e.g., "was driven" and "hit") are identified from text segments. It further outlines the extraction of associated argument roles and text (such as Vehicle, Destination, Instrument, Target, Victim). The output is a structured list of events, each comprehensively detailing its type, trigger, corresponding argument roles, and textual descriptions. This process not only enhances the accuracy of event extraction but also enriches the understanding of event contexts within the text.}
    \label{fig:2}
\end{figure}

In this study, we adopted a phased approach to optimize the event extraction process, with a particular emphasis on data balancing, as shown in Figure\ref{fig:1}.
Our task was meticulously divided into two key steps: event detection and trigger word identification, and argument role recognition. This two-step strategy not only enhances task accuracy but also provides vital contextual support for argument role recognition in the second step by effectively utilizing historical information identified in the first step. The outputs of event detection and trigger word identification directly feed into the argument role recognition stage as part of the input information, helping the model consider the overall context of events when identifying arguments, thereby establishing coherence between different stages and effectively preventing error propagation, as shown in Figure\ref{fig:2}.
In terms of data balancing, we employed a combination of undersampling and oversampling techniques to ensure representativeness of various event types in the dataset, reducing the impact of the long-tail effect on model performance. This balancing strategy is crucial for improving the model's generalization ability, especially when dealing with real-world data distributions.
\begin{figure}
    \centering
    \includegraphics[width=0.75\linewidth]{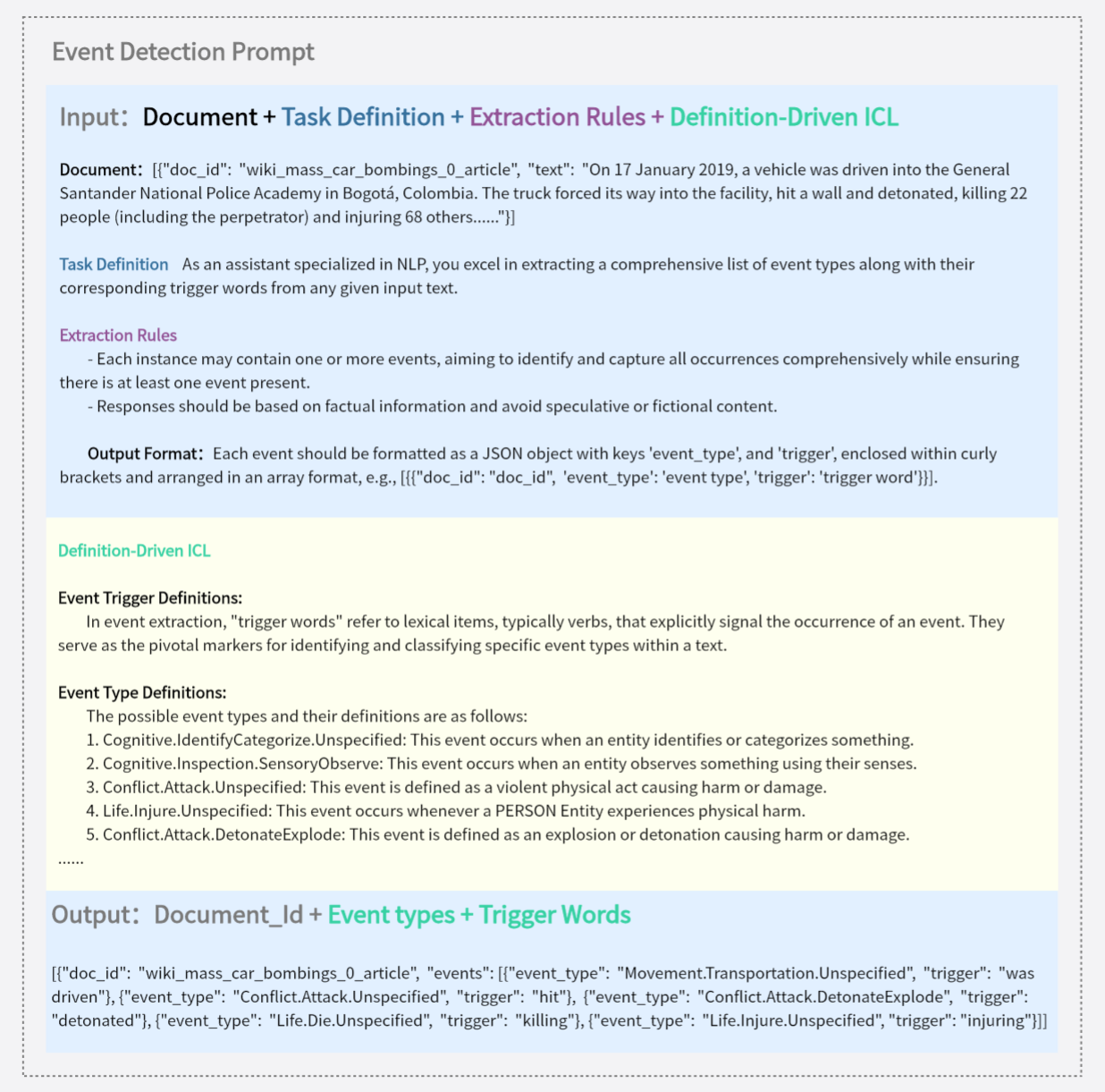}
    \caption{\textbf{Event Detection Prompting.} This figure illustrates the first step of the event extraction prompting process — event detection prompting. It involves defining the task, extracting rules, and employing Definition-Driven Interactive Constructive Learning (ICL) to extract event types and triggers from documents. The system inputs include document content, task definitions, extraction rules, and event trigger definitions. The output consists of a JSON array of objects, each containing document identifiers, event types, and triggers. For instance, it identifies event type "Movement.Transportation.Unspecified" and trigger "was driven" from the text.}
    \label{fig:3}
\end{figure}
\begin{figure}
    \centering
    \includegraphics[width=0.75\linewidth]{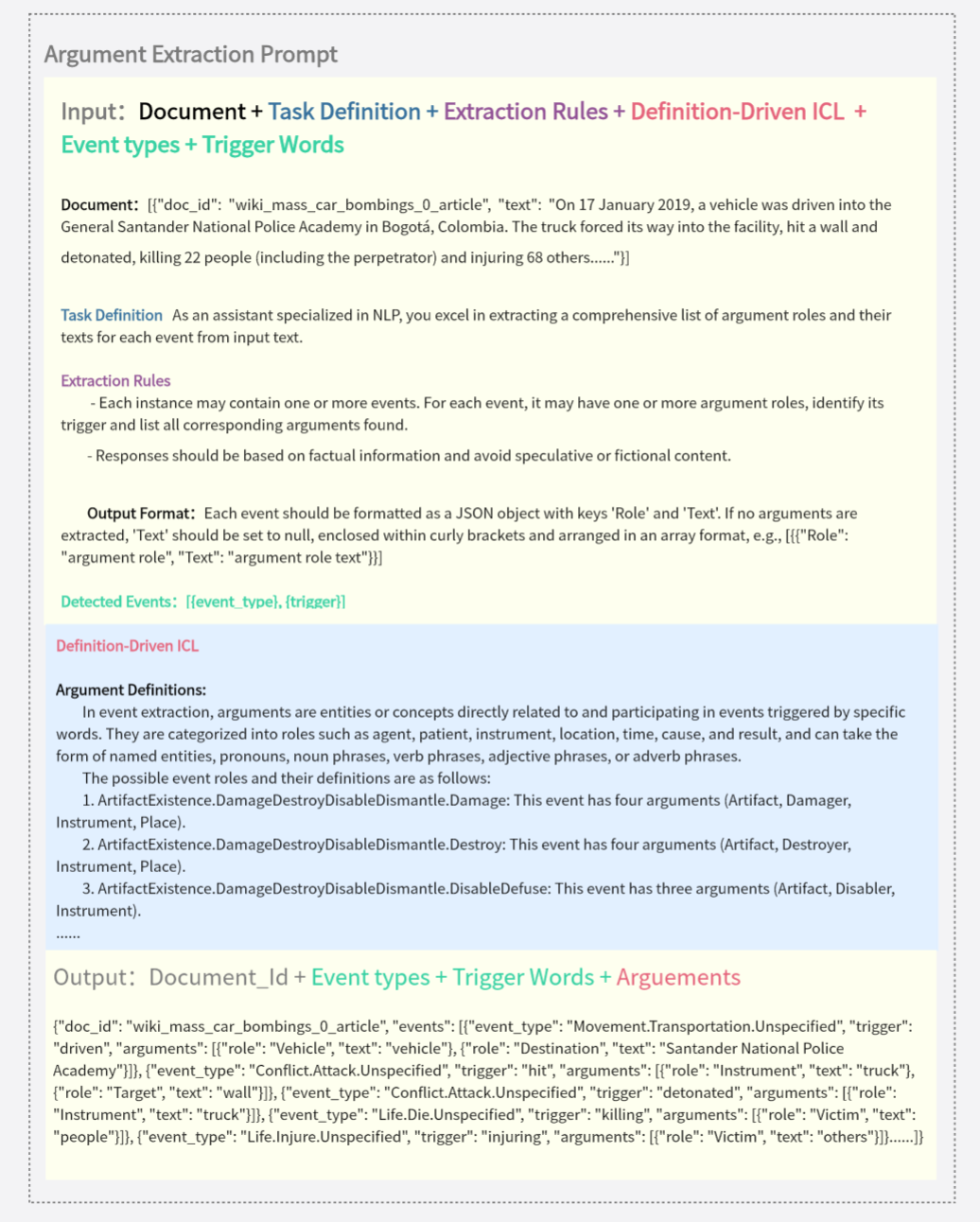}
    \caption{\textbf{Argument Role Extraction Prompting.} The figure illustrates the second step of the event extraction prompting — argument role extraction — utilizing natural language processing (NLP) techniques to extract the argument roles and their corresponding text from documents. Inputs include document content, task definitions, extraction rules, definition-driven interactive construct learning (ICL), and identified event types and triggers. The system formats each event into JSON objects following a predefined output structure, organized in array format. For instance, for the event type 'Movement.Transportation.Unspecified,' the system identifies 'driven' as the trigger, extracts 'vehicle' as the text for the 'Vehicle' role, and 'Santander National Police Academy' as the text for the 'Destination' role.}
    \label{fig:4}
\end{figure}
Regarding prompt design, we introduced an innovative heuristic approach that goes beyond traditional input-output specifications. We constructed a detailed framework for events and their argument roles, providing clear definitions, as shown in Figure\ref{fig:3} and Figure\ref{fig:4}. These definitions were automatically generated by large-scale language models using original corpora, leveraging the models' deep understanding of language and rich corpora to generate accurate and comprehensive descriptions of events and argument roles, thereby providing a rich and consistent reference baseline for the model.

Furthermore, building upon the heuristic approach, we provided chain-of-thought examples to guide the model through step-by-step reasoning. These examples not only demonstrate how to break down complex problems into more manageable sub-problems but also simulate human thought processes, providing a logical path to finding solutions. This helps the model learn how to identify key parts of problems, reason based on known information, and continuously refine and adjust its thought process during reasoning.

Through the integrated application of these methods, our aim is to enhance the performance of event extraction tasks and strengthen the model's ability to grasp complex relationships between events and arguments. This not only improves accuracy but also enhances the model's generalization and adaptability. Our research process and results demonstrate that carefully designed methods and strategies can significantly improve the efficiency and effectiveness of event extraction tasks, offering new perspectives and foundations for both the event extraction field and broader natural language processing tasks.

These innovative efforts not only bring new perspectives to the field of event extraction but also lay the foundation for future applications in a broader range of NLP tasks. Through continuous optimization and adjustment, we believe that further improvements in model performance can bring more breakthroughs to the field of natural language processing.

\section{Experiments}
\label{sec:headings}
\subsection{Dataset}
The WikiEvents dataset\cite{wiki} is a resource created to advance research in document-level event extraction, as shown in Table\ref{tab:1}. It gathers real-world event content from Wikipedia and related news reports, providing detailed textual information about events. Based on the ontology of the KAIROS project, this dataset annotates 67 event types and builds a multi-level event category system. It comprises 206 documents, 5262 sentences, and 3241 events, covering 49 event types and 57 argument types. The richness and complexity of the WikiEvents dataset offer researchers new opportunities to develop more refined and efficient event extraction models, enabling deeper understanding and processing of real-world events.

\begin{table}[htbp]
    % 插入长度为5pt的垂直空间（也可以是负数，缩进）
    \vspace{-10pt}
    \centering
    % 表名 前面为中文名/后面为英文名
    % \bicaption{总数展示表}{Total Number Display Table}
    % label标签，用以引用本表时。例：autoref{num}
    \caption{\textbf{Dataset Statistics.} This table illustrates the distribution of the Wikievents dataset, categorized into three parts: training set (Train), validation set (Dev), and test set (Test). Each part consists of varying numbers of sentences, event types, and arguments.}
    \label{tab:1}
    % 设置表格单元格的列宽
    \setlength{\tabcolsep}{10mm}{
    % 表示 三线表 有4列
    \begin{tabular}{ccccc}
    % toprule表示三线表的顶部线
        \toprule
       \multicolumn{5}{c}{\textbf{WikiEvents Dataset}}
        \\
        \midrule
        & \textbf{Documents} &
        \textbf{Sentences} & \textbf{Event Types} & \textbf{Arguments} \\
        % midrule 表示 三线表的 中部线
        \midrule

        % 合并三行1列，用空格代替，也可以用\multirow{}[]{}{}来表示
        \textbf{Train}  & 206 & 7453  & 3241    & 4542  \\
        \textbf{Dev}  & 20 & 577    & 428  & 428  \\
        \textbf{Test}  & 20 & 635    & 365   & 566  \\  
        % bottomrule表示 三线表 的底部线
        \bottomrule 
    \end{tabular}}
\end{table}

\begin{figure}
    \centering
    \includegraphics[width=0.75\linewidth]{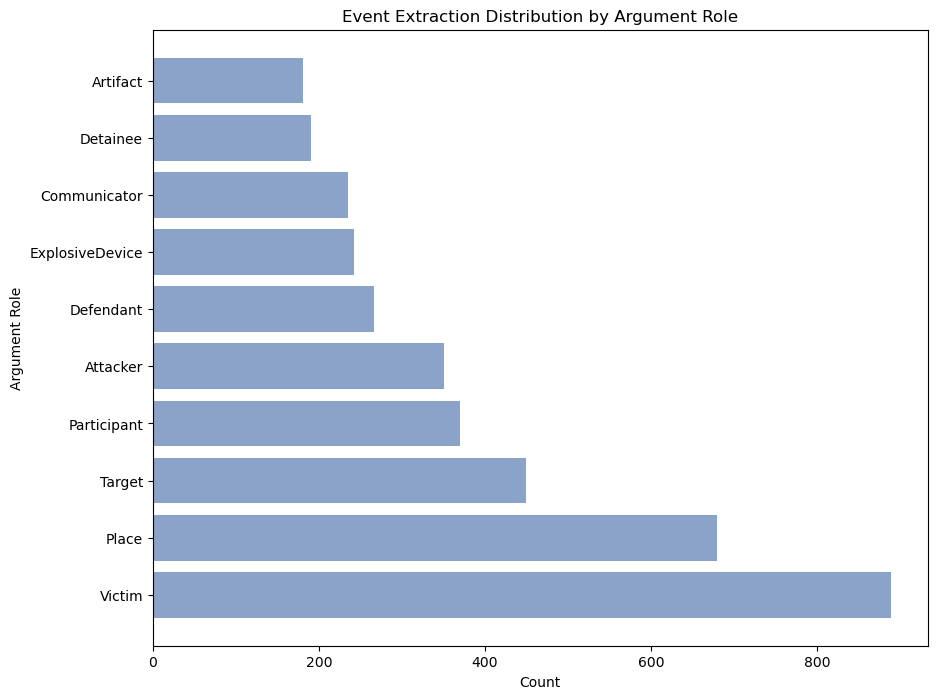}
    \caption{\textbf{Argument Role distribution.} As can be seen from the figure, the data in the original dataset are unevenly distributed with long tails.}
    \label{fig:5}
\end{figure}

Due to the long-tail distribution of event types in the data, as shown in Figure\ref{fig:5}, we conducted undersampling of the majority classes and oversampling of the minority classes to achieve data balance. Ultimately, we constructed a small dataset containing 100 samples based on 246 data points from training, validation, and test sets, ensuring roughly equal numbers of each event type.

In our experiments, we employed three large-scale language models: publicly available GPT-4, GPT-4 Turbo, and Qwen Turbo. It is noteworthy that due to the high cost associated with GPT-4, its evaluation was limited to a subset of the dataset.
\subsection{Evaluation metrics}
Building upon previous research\cite{metrix}, we adopt the following criteria to determine the correctness of predicted event mentions:

A trigger word is considered correct if its event subtype and offset match those of a reference trigger.

An argument is correctly identified if its event subtype and offset match any reference argument mention.

An argument is correctly identified and classified if its event subtype, offset, and role match any reference argument mention.

These criteria provide a systematic approach to assess the performance of event extraction models. In the context of evaluating F1 scores:

The F1 score for event detection (Trig-C) reflects the model's ability to identify event trigger words in text and to correctly classify them into specific event types. This requires not only accurate identification of triggers but also correct classification of event subtypes.

The F1 score for argument extraction (Arg-C) measures the model's capability to determine arguments associated with specific event triggers and assign them correct roles. This demands the model to accurately identify arguments and understand their roles within events.

Through these evaluation criteria and metrics, we can comprehensively assess a model's performance in event extraction tasks, including its understanding of event complexity and diversity, as well as its accuracy and reliability in practical applications. This is crucial for advancing the development and optimization of event extraction technologies.
\subsection{Baselines}
To validate our proposed method, we conducted experimental comparisons with the following event extraction models, used as baselines:
\paragraph{OntoGPT\cite{ontoGPT}.} This tool utilizes recursive querying of large language models like GPT-3, employing zero-shot learning techniques to extract knowledge from text. It applies knowledge schemas based on input texts and returns information consistent with those schemas.
\paragraph{Schema-aware Event Extraction\cite{7}.} This approach combines large language models (LLMs) with retrieval-augmented generation strategies to decompose the event extraction task into two subtasks: event detection and argument extraction. It enhances performance through customized prompts and examples, employing dynamic prompting tailored to specific query instances. It has demonstrated excellent performance across multiple benchmarks, effectively improving the accuracy and reliability of event extraction.

\section{Analysis}

\begin{table}[htbp]
    \centering
    \caption{\textbf{Event Extraction Performance.} This table summarizes the performance of various natural language processing models on the WikiEvent dataset for event extraction. The headers succinctly outline the key evaluation elements: "WikiEvent" specifies the dataset, "Model" showcases the evaluated models, and "Language model" specifies the underlying language models used by each model. The columns "Trig-C" and "Arg-C" respectively record the models" performance in trigger word recognition and classification, and argument recognition and classification, using F1 scores as a unified evaluation metric. The table includes models fine-tuned systems and unsupervised methods employing direct context learning, covering diverse technical approaches such as OntoGPT, ChatGPT, schema-aware EE, and various zero-shot methods. These models leverage language models such as "gpt-4", "ChatGPT", "Qwen-turbo", and "gpt-4-turbo".}
    \label{tab:2}
    \begin{tabular}{llcc}
        \toprule
        \textbf{Model} & \textbf{Language model} & \textbf{Trig-C} & \textbf{Arg-C} \\
        \midrule
        \multirow{2}{*}{OntoGPT\hfill\cite{ontoGPT}}& GPT-4 & 41.55 &  \textbf{29.67} \\
         & ChatGPT & 33.67 & 19.75\\
        \midrule
        \multirow{2}{*}{Schema-aware EE\hfill\cite{7}} & GPT-4 & 42.66 & 29.39 \\
         & ChatGPT & 39.08 & 24.96 \\
        \midrule
        \textbf{DDEE(Ours)}& GPT-4 & 31.47 & 24.19 \\
        
        \midrule
         \textbf{DDEE(Ours)}& Qwen-turbo & 25.93 & 20.13 \\
        \midrule
         \textbf{DDEE(Ours)}& GPT-4-turbo & \textbf{45.21} & 27.33 \\
         \midrule
         \textbf{DDEE+Cot(Ours)}& GPT-4-turbo & 11.50 & 23.78 \\
        \bottomrule
    \end{tabular}

\end{table}

\begin{figure}[t]
\centering  %图片全局居中
\subfigure[image 1]{
\label{Fig.sub.1}
\includegraphics[width=8cm,height = 8cm]{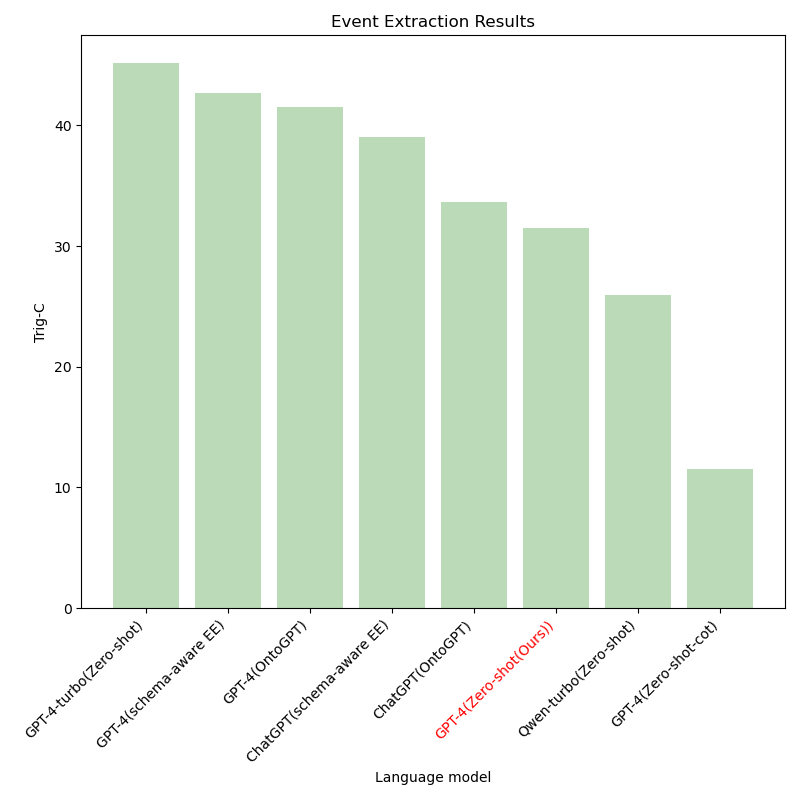}}\subfigure[image 2]{
\label{Fig.sub.2}
\includegraphics[width=8cm,height = 8cm]{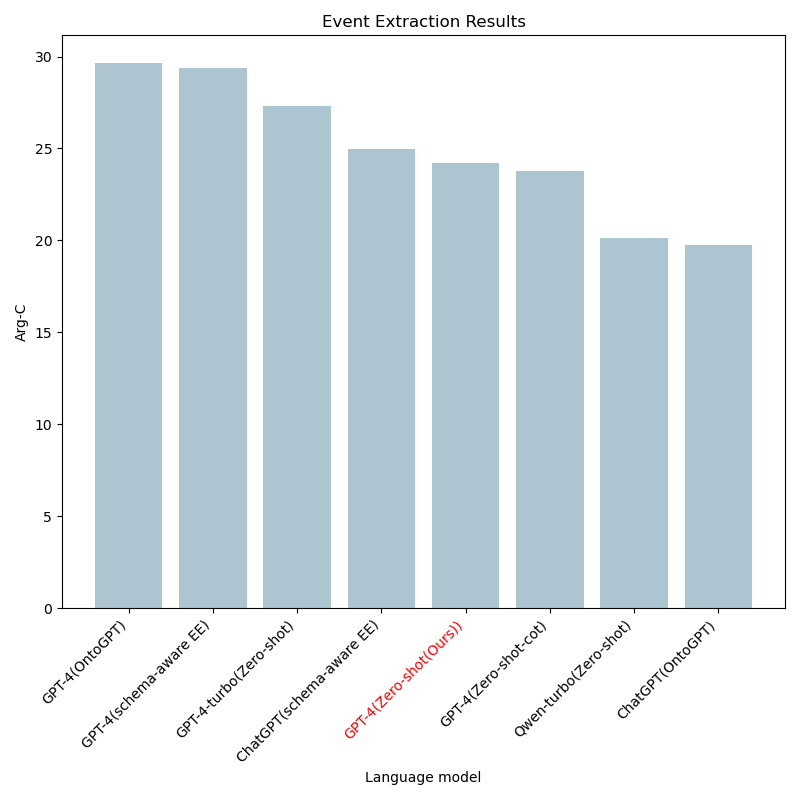}}
\caption{\textbf{Event extraction results.} The left (a)image 1 is the result of Trigger Word Recognition (Trig-C) and the (b)right image 2 is the result of Argument Role Recognition (Arg-C). Our model test results have been highlighted in red, this result shows that our model performs consistently on the task, but there is room for further optimization compared to the top models.}
\label{fig:6}
\end{figure}

% \begin{figure}
%     \centering
%     \includegraphics[width=0.5\linewidth]{}
%     \caption{Caption}
%     \label{fig:enter-label}
% \end{figure}
% \begin{figure}
%     \centering
%     \includegraphics[width=0.5\linewidth]{}
%     \caption{Caption}
%     \label{fig:enter-label}
% \end{figure}
In this study, we conducted a detailed analysis of different models' performance on the balanced WikiEvents dataset for event extraction tasks. We focused particularly on the gpt-4 language model and several zero-shot learning methods to evaluate their effectiveness in specific tasks, as shown in Table\ref{tab:2} and Figure\ref{fig:6}.

Our model (DDEE), based on gpt-4, achieved a Trigger Recognition (Trig-C) score of 31.47\% and Argument Role Recognition (Arg-C) score of 24.19\%. This result indicates that our model performs steadily on the task, but there is room for further optimization compared to top-performing models.

We also tested several LLMs  methods, including different configurations of Qwen-turbo and Gpt-4-turbo. Qwen-turbo scored 25.93\% on Trigger Recognition and 20.13\% on Argument Role Recognition. Gpt-4-turbo achieved a high score of 45.21\% on Trigger Recognition and 27.33\% on Argument Role Recognition. However, the DDEE+Cot configuration of Gpt-4 scored poorly with 11.50\% on Trigger Recognition and 23.78\% on Argument Role Recognition, indicating suboptimal performance.

It is noteworthy that while Gpt-4-turbo achieved high scores in Trigger Recognition, the performance of the DDEE+Cot configuration was below expectations. This suggests that the Cot method may not always bring significant performance improvements in non-inferential tasks, which is crucial for understanding the applicability of different learning methods in specific task types.

By introducing the balanced dataset, we effectively mitigated the impact of the long-tail effect, ensuring fairness and effectiveness in model handling of various event types. This measure is critical for enhancing model generalization, especially when dealing with real-world data distributions.

In conclusion, we believe that selecting appropriate models and configurations, coupled with well-designed prompts and balanced datasets, is crucial for improving the performance of event extraction tasks. Our research also indicates that the cot method may not be suitable for all types of tasks, especially in non-inferential tasks. Future work will focus on further optimizing model architectures, improving prompt designs, and exploring more efficient training strategies to achieve higher accuracy and efficiency.
\section{Conclusion}
This study addresses the prompt design challenges faced by large language models (LLMs) in document-level event extraction tasks and proposes an innovative optimization strategy. Through experiments, we found that adopting data balancing techniques significantly enhances the model's ability to generalize event type recognition, while finely tuned prompt designs effectively address LLMs' sensitivity to prompt styles. Additionally, the introduction of structured heuristic methods and strict constraints further improves the precision of event extraction. Our model demonstrates stable performance on the balanced WikiEvents dataset. While there is room for improvement compared to top-performing models, it has shown promising potential and application prospects.

This research not only advances event extraction technology but also provides new research perspectives and methodologies for other tasks in the NLP field. Looking ahead, we plan to apply these strategies to more diverse datasets to further validate and optimize the model's generalization and adaptability. We believe that through continuous model optimization, improvements in prompt design, and exploration of more efficient training strategies, our research will achieve higher accuracy and efficiency. Furthermore, we anticipate that these research findings will inspire innovative approaches and solutions for applying LLMs in NLP tasks, particularly in enhancing robustness and flexibility when handling diverse and complex datasets.

\section{References}
\bibliographystyle{unsrt}
\bibliography{references}

\appendix{}
\section{Appendix}
\begin{figure}
    \centering
    \includegraphics[width=0.75\linewidth]{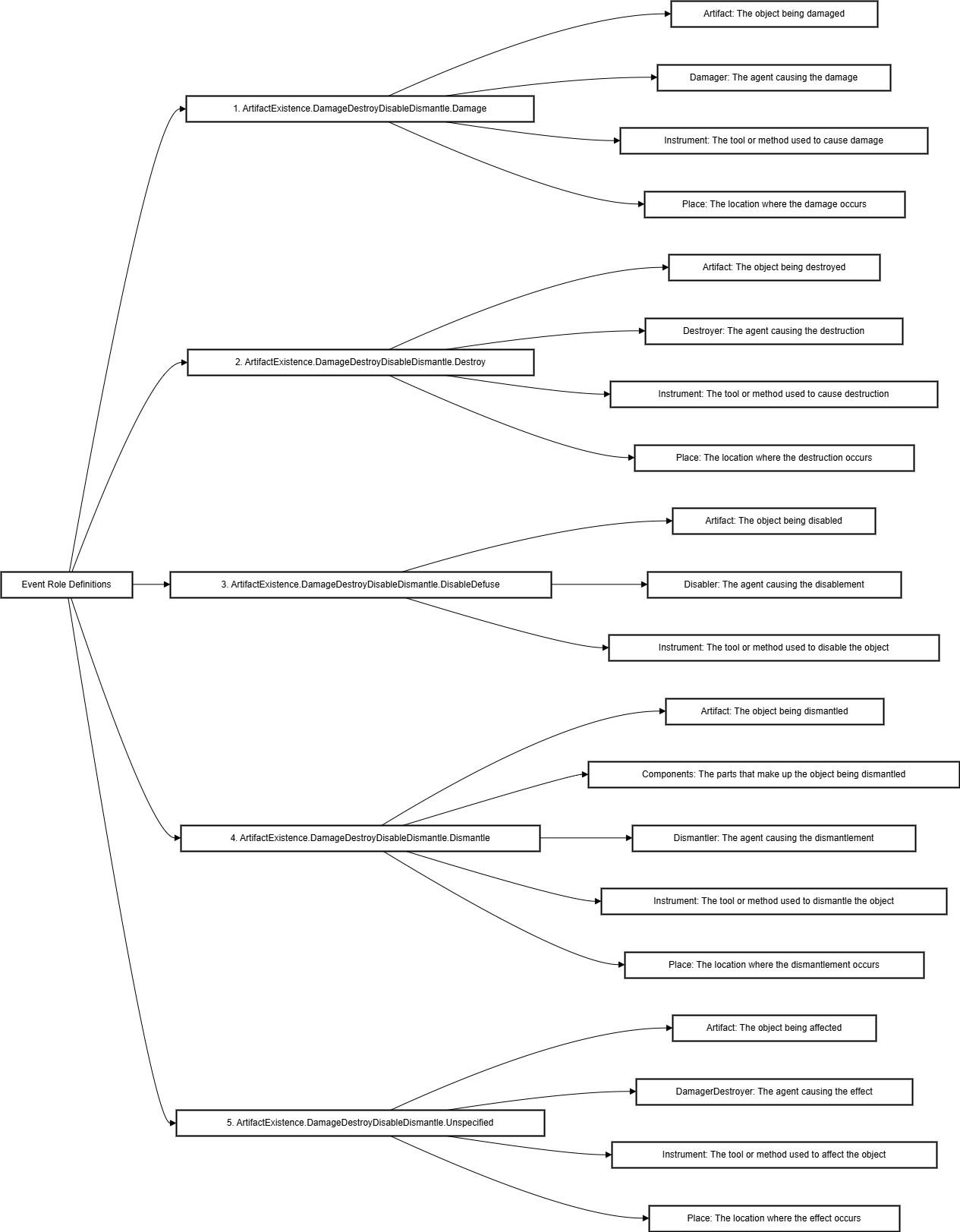}
    \caption{\textbf{Examples of WikiEvents data.} This figure shows some of the Event Types from WikiEvents and the specific definitions of their Argument Roles.}
    \label{fig:7}
\end{figure}

\end{document}